\documentclass[11pt]{article}

\usepackage[utf8]{inputenc}
\usepackage[T1]{fontenc}
\usepackage{microtype}
\usepackage{graphicx}
\usepackage{booktabs}
\usepackage{amsmath,amssymb,amsfonts}
\usepackage{algorithm}
\usepackage[noend]{algpseudocode}
\usepackage{url}
\usepackage{xcolor}
\usepackage{tikz}
\usetikzlibrary{arrows.meta,positioning,fit,backgrounds,calc}
\usepackage{pgfplots}
\pgfplotsset{compat=1.17}
\usepackage[colorlinks=true,linkcolor=blue,citecolor=blue,urlcolor=blue]{hyperref}

\usepackage[margin=1in]{geometry}
\newcommand{\icmltitlerunning}[1]{}

\definecolor{wmlblue}{HTML}{1F4E79}
\definecolor{wmlteal}{HTML}{2C7A7B}
\definecolor{wmlgray}{HTML}{E6E6E6}
\definecolor{wmlamber}{HTML}{B7791F}

\newcommand{\code}[1]{\texttt{\small #1}}

\title{\textbf{One Lens, Many Worlds\\
A Capability-Typed Interface for World-Model Interpretability}}

\author{
\normalsize\sffamily
Bhavith Chandra Challagundla\textsuperscript{1},
Sanskar Pandey\textsuperscript{2},
Param Thakkar\textsuperscript{3},
Rishikesh Mallagundla\textsuperscript{4}, \\[0.25em]
\normalsize\sffamily Yugandhar Reddy Gogireddy\textsuperscript{5},
Wenhao Lu\textsuperscript{6},
Hindol Roy Choudhury\textsuperscript{2},
Shravani Challagundla\textsuperscript{7}, \\[0.25em]
\normalsize\sffamily Mohamed Deraz Nasr\textsuperscript{8}, and Spursh Deshpande\textsuperscript{2} \\[0.9em]
{\normalfont\small\itshape\textsuperscript{1}New York University\quad \textsuperscript{2}Independent Researcher\quad \textsuperscript{3}Veermata Jijabai Technological Institute} \\[0.3em]
{\normalfont\small\itshape\textsuperscript{4}Mercity\quad \textsuperscript{5}University of Southern California\quad \textsuperscript{6}Independent Researcher, MIT} \\[0.3em]
{\normalfont\small\itshape\textsuperscript{7}Independent Researcher, GITAM\quad \textsuperscript{8}Georgia Institute of Technology}
}

\date{}

\begin{document}
\maketitle

\begin{abstract}
World models are now built on substantially different computational substrates. Latent
recurrent state-space models such as PlaNet and the Dreamer family compress observations into
recurrent states; token-based models such as IRIS quantize observations into a learned codebook
and predict autoregressively with a transformer; and joint-embedding predictive architectures
such as I-JEPA predict in a learned latent space with no pixel decoder. The interpretability
methods applied to these models, including probing, activation patching, sparse autoencoders,
and surprise analysis, share a common set of primitives, yet they are re-implemented from
scratch for each architecture because existing hook-and-cache tooling assumes a transformer
language model with no notion of actions, environment steps, or imagined rollouts. We argue
that this fragmentation reflects the tooling rather than the models, and that the shared
structure of world models is captured by a small typed interface. We present WorldModelLens, an
open-source interpretability substrate organized around a capability-typed adapter: every model
implements four required methods (encode, transition, initial state, sample) and declares a set
of optional heads (decode, reward, continue, actor, critic) through an explicit capability
descriptor, so that reinforcement-learning and self-supervised world models are first-class
without either imitating the other. A single hook and cache layer exposes time-indexed
activations, imagination rollouts, and intervention replay over this interface, allowing each
analysis to be written once. We demonstrate the full analysis suite end-to-end on I-JEPA,
provide adapters spanning latent-recurrent, transformer-token, and joint-embedding families as
evidence that the interface generalizes by construction, and show that the hook layer adds
approximately 12\% per-step overhead when inactive, which makes always-on interpretability
telemetry practical inside training and control loops.
\end{abstract}

\section{Introduction}
\label{sec:intro}

World models have become a central tool for learning predictive representations of environments
\cite{ha2018worldmodels,hafner2019planet}, and the architectures used to build them have
diverged sharply. The Dreamer family and PlaNet learn a recurrent state-space model whose latent
combines a deterministic recurrent component with a continuous or discrete-categorical stochastic
component \cite{hafner2020dreamer,hafner2021dreamerv2,hafner2023dreamerv3}. IRIS and Decision
Transformer quantize observations into a learned codebook and model dynamics autoregressively
with a transformer \cite{micheli2023iris,chen2021decisiontransformer}. Joint-embedding predictive
architectures such as I-JEPA and TD-MPC2 predict directly in a learned embedding space and may
have no decoder at all \cite{assran2023ijepa,hansen2024tdmpc2}. These models differ not only in
their internal computation but in their interface to the environment: some consume actions and
emit rewards and values, while others are purely self-supervised and have neither.

Interpretability methods, by contrast, are uniform in spirit. Linear and nonlinear probes read an
internal representation and predict a property of interest \cite{alain2016probes,belinkov2022probing};
activation patching overwrites an internal activation and measures the downstream effect
\cite{meng2022rome,wang2023ioi}; sparse autoencoders decompose activations into interpretable
features \cite{cunningham2024saes,bricken2023monosemanticity}. Each reduces to the same three
primitives: read an activation, optionally overwrite it, and observe an effect. This regularity
is exactly what mechanistic-interpretability tooling for language models exploits. TransformerLens
\cite{nanda2022transformerlens} popularized a hook-and-cache abstraction in which named hook points
expose every internal activation for reading or editing, and a body of work has built circuit-level
analyses on top of it \cite{elhage2021mathematical,wang2023ioi}.

That tooling, however, is shaped around the transformer language model. It assumes a stack of
attention blocks operating on a sequence of tokens, with no representation for an action input, no
environment step, no multi-step rollout, no imagined future, and no mixed continuous and discrete
latent. World models violate every one of these assumptions. The practical effect is that
interpretability code is rewritten for each new world model, results are reported in
model-specific terms that do not transfer, and questions that depend on comparing internal
structure across architectures cannot be expressed in shared code. General-purpose interpretability
libraries such as Captum \cite{kokhlikyan2020captum}, NNsight \cite{fiottokaufman2024nnsight}, and
Penzai \cite{johnson2024penzai} provide attribution and access primitives but do not model
world-model semantics such as trajectories, dynamics rollouts, or intervention replay.

We argue that this fragmentation is a property of the tooling rather than of the models. A world
model, irrespective of family, can be described by an encoder that maps an observation to a latent,
a transition that advances latent state and optionally consumes an action, and a set of optional
read-out heads that a given model may or may not possess. We make this structure explicit through a
\emph{capability-typed adapter}, in which the required core must be implemented and optional
capabilities are declared by a descriptor, so that a reinforcement-learning agent and a
self-supervised predictor are both first-class without either having to fabricate the other's heads.
A single hook and cache layer mounted over this interface then exposes time-indexed activations,
imagination rollouts, and intervention replay, so that every analysis is written once against the
interface rather than against any single architecture.

We present WorldModelLens, an open-source realization of this design. Our contributions are:
\begin{itemize}
\setlength{\itemsep}{1pt}
\item A capability-typed world-model interface (Section~\ref{sec:abstraction}) that expresses
reinforcement-learning and self-supervised world models uniformly through four required methods and
five optional, descriptor-gated heads, and that we instantiate across the Dreamer, PlaNet, TD-MPC2,
IRIS, Decision Transformer, and I-JEPA families.
\item A backend-agnostic hook, cache, rollout, and intervention-replay layer over that interface,
with time-indexed activation caching and device offloading for long rollouts, on top of which
probing, activation patching, sparse autoencoders, and surprise analysis are each implemented
exactly once (Section~\ref{sec:analyses}).
\item An end-to-end demonstration on I-JEPA (Section~\ref{sec:eval}) in which the unmodified
analysis suite recovers layer-resolved structure in the predictor, together with a measurement of
the overhead the substrate imposes.
\item An open implementation whose hook layer adds approximately 12\% per-step overhead when
inactive, making always-on interpretability telemetry practical inside training and control loops.
\end{itemize}
Scaling the empirical analysis to additional families, including Dreamer at full scale, V-JEPA
\cite{bardes2024vjepa}, and Cosmos \cite{nvidia2025cosmos}, is in progress and forms our roadmap
(Section~\ref{sec:limitations}). The present paper establishes the interface, the substrate, and a
deep single-family demonstration.

\section{Background}
\label{sec:background}

We formalize a world model as a tuple
\begin{equation}
\mathcal{M} = \big(\mathcal{O},\, \mathcal{S},\, \mathcal{Z},\, \mathcal{A},\ \iota,\, \mathcal{E},\, \tau,\ \mathcal{H}\big),
\end{equation}
where $\mathcal{O}$ is the observation space, $\mathcal{S}$ the deterministic latent state space,
$\mathcal{Z}$ the stochastic latent space, and $\mathcal{A}$ an optional action space. The required
core is three maps and a sampler. An initial-state map $\iota:\{\ast\}\to\mathcal{S}$ returns $s_0$.
A probabilistic encoder
\begin{equation}
\mathcal{E}:\mathcal{O}\times\mathcal{S}\to\Delta(\mathcal{Z}),
\qquad z_t \sim q_\theta(z \mid o_t, s_{t-1})
\end{equation}
maps an observation and the previous state to a distribution over latent codes in the variational tradition \cite{kingma2014vae}, from which
$\textsc{sample\_z}$ draws $z_t$ using a Gumbel-softmax relaxation \cite{jang2017gumbel,maddison2017concrete}
when $\mathcal{Z}$ is categorical and the identity when it is continuous. A transition
\begin{equation}
\tau:\mathcal{S}\times\mathcal{Z}\times(\mathcal{A}\cup\{\varnothing\})\to\mathcal{S},
\qquad s_{t+1} = \tau(s_t, z_t, a_t),
\end{equation}
advances the deterministic state, with $a_t=\varnothing$ for action-free models. The unit visible to
downstream analysis is the joint latent $h_t = (s_t, z_t)$. The optional heads form a set
$\mathcal{H}\subseteq\{g_{\mathrm{dec}}, g_{\mathrm{rew}}, g_{\mathrm{cont}}, \pi, V\}$ with signatures
\begin{equation}
\begin{aligned}
g_{\mathrm{dec}}&:\mathcal{S}\!\times\!\mathcal{Z}\to\mathcal{O}, &
g_{\mathrm{rew}}&:\mathcal{S}\!\times\!\mathcal{Z}\to\mathbb{R}, &
g_{\mathrm{cont}}&:\mathcal{S}\!\times\!\mathcal{Z}\to[0,1], \\
\pi&:\mathcal{S}\!\times\!\mathcal{Z}\to\Delta(\mathcal{A}), &
V&:\mathcal{S}\!\times\!\mathcal{Z}\to\mathbb{R}, &&
\end{aligned}
\end{equation}
for decoding, reward, continuation, policy, and value. Reinforcement-learning world models
\cite{sutton2018rl} instantiate most of $\mathcal{H}$; self-supervised video and joint-embedding models instantiate few
or none. A rollout over $o_{1:T}$ is the recursion $z_t \sim \mathcal{E}(o_t, s_{t-1})$,
$s_{t+1}=\tau(s_t,z_t,a_t)$, and \emph{imagination} is the same recursion with $\mathcal{E}$ replaced
by the learned prior $p_\theta(z\mid s_t)$. The per-step \emph{surprise} is the divergence between the
posterior and prior latents,
\begin{equation}
\label{eq:surprise}
\mathrm{surprise}_t \;=\; D_{\mathrm{KL}}\!\big(q_\theta(z\mid o_t, s_{t-1})\,\big\|\,p_\theta(z\mid s_t)\big).
\end{equation}

The families we target differ chiefly in the form of the latent and the transition, not in this
signature. Dreamer and PlaNet use a recurrent state-space model whose latent concatenates a
deterministic recurrent part with a stochastic part that is continuous in V1 and
discrete-categorical from V2 onward \cite{hafner2020dreamer,hafner2021dreamerv2}. IRIS and Decision
Transformer use a transformer \cite{vaswani2017attention} over a discrete codebook, so the latent is a sequence of tokens
\cite{micheli2023iris,chen2021decisiontransformer}. I-JEPA and TD-MPC2 predict in a continuous
embedding space and, in the case of I-JEPA, have no decoder \cite{assran2023ijepa,hansen2024tdmpc2}.
A transformer-only hook library has no place for the action input, no representation for an
imagination rollout, and no abstraction that covers a recurrent latent, a token sequence, and a
joint embedding at once. The interface above does, and it is this interface that WorldModelLens
makes concrete.

\section{The WorldModelLens Abstraction}
\label{sec:abstraction}

\begin{figure}[t]
\centering
\begin{tikzpicture}[
  font=\footnotesize,
  comp/.style={draw=wmlteal, thick, rounded corners=2pt, fill=white,
               minimum width=2.55cm, minimum height=0.6cm, align=center, font=\scriptsize},
  arr/.style={-{Stealth[length=2.6mm]}, very thick, wmlblue},
  band/.style={draw=wmlblue, very thick, rounded corners=4pt},
  title/.style={font=\footnotesize\bfseries, text=wmlblue}
]
\def\ya{4.7}\def\yb{2.35}\def\yc{0.0}
\node[comp] at (-2.95,\ya) {Probing};
\node[comp] at (0,\ya) {Patching};
\node[comp] at (2.95,\ya) {SAE / Surprise};
\node[comp] at (-2.95,\yb) {run\_with\_cache};
\node[comp] at (0,\yb) {run\_with\_hooks};
\node[comp] at (2.95,\yb) {imagine / replay};
\node[comp] at (-2.95,\yc) {Dreamer / PlaNet};
\node[comp] at (0,\yc) {IRIS / DT};
\node[comp] at (2.95,\yc) {I-JEPA / TD-MPC2};
\begin{scope}[on background layer]
  \fill[wmlgray]     (-4.6,\ya-0.5) rectangle (4.6,\ya+1.0);
  \fill[wmlteal!12]  (-4.6,\yb-0.5) rectangle (4.6,\yb+1.0);
  \fill[wmlamber!14] (-4.6,\yc-0.5) rectangle (4.6,\yc+1.0);
  \draw[band] (-4.6,\ya-0.5) rectangle (4.6,\ya+1.0);
  \draw[band] (-4.6,\yb-0.5) rectangle (4.6,\yb+1.0);
  \draw[band] (-4.6,\yc-0.5) rectangle (4.6,\yc+1.0);
\end{scope}
\node[title] at (0,\ya+0.72) {Analysis layer (written once)};
\node[title] at (0,\yb+0.72) {Hook \& cache layer: HookedWorldModel};
\node[title] at (0,\yc+0.72) {Adapter layer: BaseModelAdapter \(+\) Capabilities};
\draw[arr] (-3.9,\yb+1.0) -- (-3.9,\ya-0.5);
\draw[arr] (-3.9,\yc+1.0) -- (-3.9,\yb-0.5);
\draw[arr] ( 3.9,\yb+1.0) -- ( 3.9,\ya-0.5);
\draw[arr] ( 3.9,\yc+1.0) -- ( 3.9,\yb-0.5);
\node[font=\scriptsize\itshape, text=wmlblue] at (0,\ya-0.78) {activations read and edited by name};
\node[font=\scriptsize\itshape, text=wmlblue] at (0,\yb-0.78) {required \(+\) optional model interface};
\end{tikzpicture}
\caption{The three-layer design. Adapters expose any world model through one typed interface; the
hooked-model wrapper mounts named hook points and a single cache; analyses read and edit activations
without referring to any architecture.}
\label{fig:overview}
\end{figure}
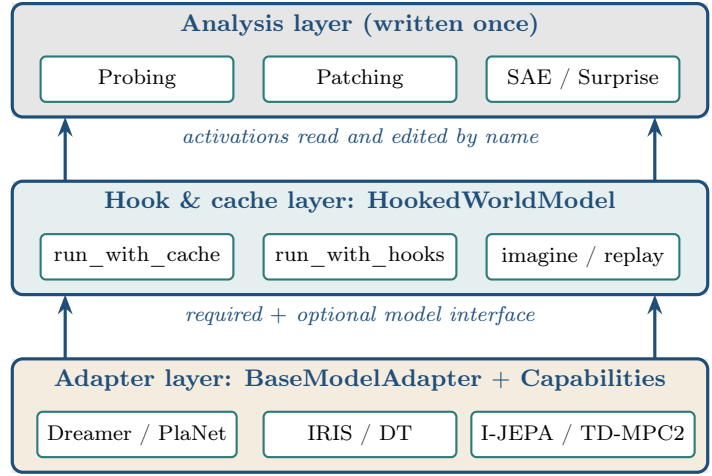

WorldModelLens is organized as three layers, shown in Figure~\ref{fig:overview}, that separate what
a model \emph{is} from how it is \emph{instrumented} from what is \emph{measured}. At the bottom, a
backend adapter exposes a single world model through the required-and-optional interface of
Section~\ref{sec:background}, translating that model's particular internals into the common
signature. In the middle, the \code{HookedWorldModel} wrapper mounts named hook points over the
adapter's outputs and routes every activation it produces through a single cache manager, so that
each activation can be read, recorded, or overwritten by name. At the top, a library of analyses
operates entirely through the wrapper, reading and editing activations by name and never referring
to a particular architecture.

The value of this separation is that responsibility is partitioned cleanly across the boundaries. A
forward pass enters at the adapter, which calls \code{encode} on each observation and
\code{transition} to advance the latent state, emitting a named activation at every hook point. The
wrapper records or modifies those activations and assembles them into a typed trajectory. The
analysis layer then consumes the trajectory, or installs interventions that the wrapper replays
through the adapter. Because the adapter is the only layer that knows how a given model computes,
replacing DreamerV3 with I-JEPA changes only the bottom layer, while every analysis above it
continues to run without modification. This is the concrete mechanism behind the portability claim
of Section~\ref{sec:intro}: an analysis is written against the interface once, and inherits every
present and future backend for free. The remainder of this section describes each layer in turn.

\subsection{Capability-Typed Adapters}
\label{sec:adapters}

A backend is registered by subclassing \code{BaseModelAdapter}, a \code{torch.nn.Module} whose
required surface is four methods: \code{encode(obs, h\_prev)}, which returns a posterior latent and
its prior; \code{transition(h, z, action)}, which advances the deterministic state and accepts an
optional action; \code{initial\_state(batch\_size, device)}; and \code{sample\_z(logits, temperature)},
which draws a latent from the encoder output and applies a Gumbel-softmax relaxation
\cite{jang2017gumbel,maddison2017concrete} for categorical latents and the identity for continuous
ones. Five further methods are optional and correspond to read-out heads: \code{decode},
\code{predict\_reward}, \code{predict\_continue}, \code{actor\_forward}, and \code{critic\_forward}.
Each optional method raises \code{NotImplementedError} in the base class, so a head is available only
when an adapter overrides it. Formally, the declared capabilities form a vector
$c\in\{0,1\}^{7}$ over $(\textsc{dec},\textsc{rew},\textsc{cont},\textsc{act},\textsc{crit},\textsc{usesA},\textsc{rl})$,
normalized on construction to satisfy the implications
\begin{equation}
c_{\textsc{act}} \Rightarrow c_{\textsc{usesA}}, \qquad c_{\textsc{rew}} \Rightarrow c_{\textsc{rl}},
\end{equation}
and the instantiated head set is $\mathcal{H}(c)=\{g_j : c_j = 1\}$. An analysis dispatches on
$\mathcal{H}(c)$ rather than on the concrete model class, which is what lets a single implementation
serve reinforcement-learning and self-supervised models alike.

\begin{figure}[t]
\centering
\begin{tikzpicture}[
  font=\scriptsize,
  core/.style={draw=wmlblue, thick, fill=wmlblue!12, rounded corners=2pt,
               minimum width=2.2cm, minimum height=0.78cm, align=center},
  on/.style={draw=wmlteal, thick, fill=wmlteal!18, rounded corners=2pt,
             minimum width=2.2cm, minimum height=0.56cm, align=center},
  off/.style={draw=gray!55, dashed, fill=white, rounded corners=2pt,
              minimum width=2.2cm, minimum height=0.56cm, align=center, text=gray!60},
  hdr/.style={font=\footnotesize\bfseries, text=wmlblue}
]
\node[hdr, align=right] at (-2.85,0.7)  {Required\\core};
\node[hdr, align=right] at (-2.85,-1.45) {Optional\\heads};

\node[hdr] at (0,1.75)  {DreamerV3};
\node[hdr] at (2.7,1.75) {IRIS};
\node[hdr] at (5.4,1.75) {I-JEPA};

\foreach \x in {0,2.7,5.4}{ \node[core] at (\x,0.7) {encode, transition,\\ init, sample}; }

\node[on]  at (0,-0.65)  {decode};
\node[on]  at (0,-1.30)  {reward, continue};
\node[on]  at (0,-1.95)  {actor, critic};

\node[on]  at (2.7,-0.65) {decode (tokens)};
\node[off] at (2.7,-1.30) {reward};
\node[off] at (2.7,-1.95) {actor, critic};

\node[off] at (5.4,-0.65) {decode};
\node[off] at (5.4,-1.30) {reward, continue};
\node[off] at (5.4,-1.95) {actor, critic};

\begin{scope}[on background layer]
\draw[draw=gray!40, rounded corners=4pt] (-3.95,-2.45) rectangle (6.65,1.45);
\draw[gray!35, dashed] (-3.95,-0.05) -- (6.65,-0.05);
\end{scope}
\end{tikzpicture}
\caption{The same interface, populated differently. Solid teal boxes are heads a family exposes;
dashed gray boxes are capabilities it declares absent. The required core (blue) is identical across
all families, so analyses target it directly and consult the descriptor for everything optional.}
\label{fig:capabilities}
\end{figure}
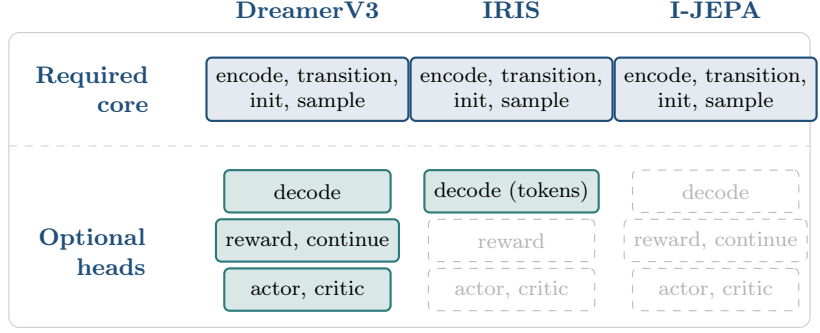

Which optional heads exist is declared rather than discovered. A \code{WorldModelCapabilities}
descriptor carries seven boolean fields, \code{has\_decoder}, \code{has\_reward\_head},
\code{has\_continue\_head}, \code{has\_actor}, \code{has\_critic}, \code{uses\_actions}, and
\code{is\_rl\_trained}, and normalizes them on construction: declaring an actor implies the model is
action-conditioned, and declaring a reward head implies it was trained against a
reinforcement-learning objective. Two predicates, \code{requires\_actions} and \code{is\_rl\_model},
summarize the descriptor for callers. Analyses consult these fields and skip what a model does not
provide instead of failing on a missing method. Figure~\ref{fig:capabilities} shows three families
populating the same interface differently: an I-JEPA encoder declares no reward, value, or action
capability, a planning model such as PlaNet declares a decoder and a reward head but no actor, and a
DreamerV3 agent declares the full set. Adapters are resolved through a central \code{BackendRegistry}
indexed by a \code{WorldModelFamily} enumeration and by capability, so a backend is selected by family
or by the features an analysis requires rather than by a hard-coded class reference.

\subsection{Hooks, Caching, and Time}
\label{sec:hooks}

The methods of an adapter are wrapped by \code{HookedWorldModel}, which mounts named hook points and
routes every activation through a single \code{HookCacheManager}. A hook point is identified by a
component name and a time index, and a hook function receives the tensor at that point together with
a \code{HookContext} that carries the current timestep, the component name, and the trajectory
observed so far. Hooks run at a \code{pre} or \code{post} stage relative to the component and can be
restricted to a single timestep or a half-open range of timesteps, which gives temporal control
without altering the model. A higher-level hook grammar parses string specifications such as
\code{z[0:10]} for a range of latent dimensions, \code{t=5.z} for a single step, and
\code{transition.pre} for a stage, and compiles them down to the same primitives. For models exposed
as ordinary \code{nn.Module} graphs, a \code{HookedRootModule} registers forward hooks on leaf
modules and assigns standardized names such as \code{encoder.layer\_i.hook\_output} and
\code{attn.hook\_\{query,key,value,pattern\}}, which makes the attention internals of
transformer-token and joint-embedding models addressable by the same machinery.

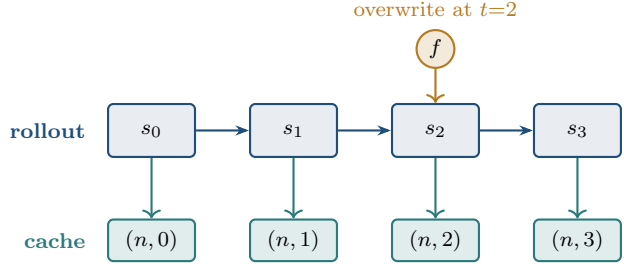
\begin{figure}[t]
\centering
\begin{tikzpicture}[
  font=\scriptsize,
  wstate/.style={draw=wmlblue, thick, fill=wmlblue!10, rounded corners=2pt,
               minimum width=1.15cm, minimum height=0.7cm, align=center},
  cache/.style={draw=wmlteal, thick, fill=wmlteal!14, rounded corners=2pt,
                minimum width=1.15cm, minimum height=0.55cm, align=center},
  hook/.style={draw=wmlamber, thick, fill=wmlamber!16, circle, inner sep=1pt, minimum size=0.5cm},
  flow/.style={-{Stealth[length=1.8mm]}, thick, wmlblue}
]
\node[wstate] (s0) {$s_0$};
\node[wstate, right=0.7cm of s0] (s1) {$s_1$};
\node[wstate, right=0.7cm of s1] (s2) {$s_2$};
\node[wstate, right=0.7cm of s2] (s3) {$s_3$};
\draw[flow] (s0)--(s1); \draw[flow] (s1)--(s2); \draw[flow] (s2)--(s3);

\foreach \i in {0,1,2,3}{ \node[cache, below=0.8cm of s\i] (c\i) {$(n,\i)$}; \draw[->, thick, wmlteal] (s\i)--(c\i); }
\node[left=0.15cm of c0, text=wmlteal, font=\scriptsize\bfseries] {cache};
\node[left=0.15cm of s0, text=wmlblue, font=\scriptsize\bfseries] {rollout};

\node[hook, above=0.45cm of s2] (hk) {$f$};
\draw[->, thick, wmlamber] (hk) -- (s2);
\node[above=0.04cm of hk, text=wmlamber, font=\scriptsize] {overwrite at $t{=}2$};
\end{tikzpicture}
\caption{Two access patterns over a rollout. \code{run\_with\_cache} (teal) records every activation
under a (name, $t$) key for later analysis. \code{run\_with\_hooks} (amber) installs a function $f$
that overwrites a chosen activation in place, which is the basis of patching, ablation, and
intervention replay.}
\label{fig:dataflow}
\end{figure}

Two entry points cover the analysis surface. \code{run\_with\_cache} performs a full forward pass and
writes every requested activation into a time-indexed \code{ActivationCache} addressable by the pair
(name, $t$); the cache stores tensors, lazily evaluated callables, and \code{torch.distributions}
objects, the last of which lets it compute per-step surprise as the Kullback-Leibler divergence
between the posterior and prior latent distributions \cite{hafner2021dreamerv2}.
\code{run\_with\_hooks} installs temporary functions at named points for the duration of one pass,
which is the primitive underlying activation patching \cite{meng2022rome,wang2023ioi}, ablation, and
intervention. Writing $\mathcal{C}$ for the cache as a partial map $(n,t)\mapsto\mathcal{C}[n,t]$ from
a hook name and timestep to a tensor, \code{run\_with\_hooks} replaces the activation at a chosen site
by a function $\phi_{n,t}$ before it propagates, $h_{n,t}\leftarrow\phi_{n,t}(h_{n,t})$, which subsumes
ablation ($\phi\equiv 0$ on a coordinate subset), additive noising
($\phi(h)=h+\epsilon$, $\epsilon\sim\mathcal{N}(0,\sigma^2 I)$), and restoration
($\phi\equiv\mathcal{C}^{\mathrm{clean}}[n,t]$ from a cached clean run). Long rollouts can exceed
device memory, so the cache exposes an offloading policy
(retain on device, move to host, spill to disk, or switch automatically past a timestep threshold)
with optional half-precision quantization, while a lazy trajectory store backs episodes with shared
tensors persisted in Zarr or HDF5. Figure~\ref{fig:dataflow} contrasts the two entry points.

\subsection{Trajectories, Imagination, and Intervention Replay}
\label{sec:trajectories}

Above the cache sits a typed data model. A \code{WorldState} records one step as a required latent
state with optional action, reward, value, and termination fields, and a \code{WorldTrajectory} is a
sequence of such states tagged as real, imagined, or planned. A parallel \code{LatentState} and
\code{LatentTrajectory} expose the recurrent and stochastic components of latent-dynamics models
directly, including the per-step Kullback-Leibler term used as a surprise signal. Two operations
distinguish world-model interpretability from its language-model counterpart. Imagination, exposed
as \code{imagine}, rolls the transition forward from a chosen state without further observations and
samples actions from the actor when one is present, which produces counterfactual futures that are
themselves cached and analyzed. Intervention replay re-executes a recorded trajectory while
installing hooks that overwrite chosen activations, so the causal effect of an internal edit on the
model's own predicted future is measured directly rather than inferred. A three-level hierarchy ties
these together at the granularity of a dimension, a state, or a trajectory, so an analysis can locate
a behavior over a span of steps, examine the active dimensions of a single state, and trace a result
down to one latent coordinate. Together these convert the static read-and-patch primitives of
language-model interpretability \cite{nanda2022transformerlens} into the rollout-and-replay
primitives that world-model analysis requires. These three layers, the adapter that fixes the
interface, the hooked wrapper that fixes how it is observed, and the typed trajectories that fix what
is recorded, are the substrate against which every analysis in the next section is written.

\section{Analyses, Expressed Once}
\label{sec:analyses}

Because every adapter exposes the same interface, an analysis written against that interface runs on
every backend without modification. Algorithm~\ref{alg:patch} states activation patching in terms of
\code{run\_with\_cache} and \code{run\_with\_hooks} alone. A clean run is cached, a corrupted run is
patched at one site with the cached value, and the reported effect is the recovery rate
$(m_{\text{patched}} - m_{\text{corrupted}}) / (m_{\text{clean}} - m_{\text{corrupted}})$, clamped to
the unit interval. The same two primitives support causal tracing that adds Gaussian noise to all
components and restores one at a time \cite{meng2022rome,geiger2021causal}, and a greedy circuit search that ablates
attention heads in order of attributed effect and keeps the minimal set that preserves a behavior
\cite{conmy2023acdc,goldowskydill2023pathpatching}.

The remaining analyses follow the same read-or-edit pattern over named activations and never
reference an architecture. Probing fits linear, ridge, and logistic estimators with stratified
cross-validation and reports permutation-test significance
\cite{alain2016probes,belinkov2022probing,hewitt2019designing}. Semantic probes project latents onto
frozen DINO \cite{caron2021dino,oquab2024dinov2} and CLIP \cite{radford2021clip} features, and a learned linear map
supports natural-language concept queries against the latent space \cite{kim2018tcav}. Sparse autoencoders are provided
in ReLU, top-$k$ \cite{gao2024topk}, and gated \cite{rajamanoharan2024gated} forms, trained with a
reconstruction-plus-sparsity objective in the tradition of dictionary learning
\cite{olshausen1996sparse,cunningham2024saes,bricken2023monosemanticity,templeton2024scaling}, and the evaluator reports
the expected $L_0$, reconstruction fidelity, and dead-feature fraction. Disentanglement is measured
with the mutual-information gap, the DCI scores, and SAP
\cite{chen2018mig,eastwood2018dci,kumar2018sap}; representational similarity uses centered kernel
alignment \cite{kornblith2019cka}; faithfulness is scored by the area over the perturbation curve
\cite{samek2017aopc}; attribution combines integrated gradients \cite{sundararajan2017ig} with
SmoothGrad \cite{smilkov2017smoothgrad}; and uncertainty is decomposed into epistemic and aleatoric
parts, with a Mahalanobis score for out-of-distribution latents \cite{lee2018mahalanobis}. Each is
implemented once and inherits every backend.

For precision, the principal estimands are the sparse-autoencoder objective, integrated-gradients
attribution, and centered kernel alignment,
\begin{align}
\mathcal{L}_{\mathrm{SAE}}(x) &= \lVert x - W_d f\rVert_2^2 + \lambda\lVert f\rVert_1, \quad f=\sigma_k(W_e x + b), \label{eq:sae}\\
\mathrm{IG}_i(x) &= (x_i - x_i')\int_0^1 \frac{\partial F\!\big(x' + \alpha(x-x')\big)}{\partial x_i}\,d\alpha, \label{eq:ig}\\
\mathrm{CKA}(X,Y) &= \frac{\mathrm{HSIC}(K,L)}{\sqrt{\mathrm{HSIC}(K,K)\,\mathrm{HSIC}(L,L)}}, \label{eq:cka}
\end{align}
where $\sigma_k$ is the top-$k$ activation, $W_e, W_d$ the encoder and decoder weights, $x'$ a
baseline embedding, and $K,L$ the Gram matrices of the two representations; the integral in
\eqref{eq:ig} is approximated by a $50$-step Riemann sum. Faithfulness is the area over the
perturbation curve, $\mathrm{AOPC}=\tfrac{1}{K+1}\sum_{k=0}^{K}\!\big(m_0 - m_k\big)$, where $m_k$ is
the prediction after ablating the $k$ most important coordinates, and an out-of-distribution latent
is scored by the Mahalanobis distance $d_M(z)=\sqrt{(z-\mu)^\top\Sigma^{-1}(z-\mu)}$ to the fitted
latent Gaussian $\mathcal{N}(\mu,\Sigma)$.

\begin{algorithm}[t]
\caption{Activation patching, written once against the interface}
\label{alg:patch}
\begin{algorithmic}[1]
\State \textbf{Input:} hooked model $M$, clean obs $o$, corrupted obs $o'$, hook name $n$, step $t$
\State $\_,\ \mathcal{C} \gets M.\textsc{run\_with\_cache}(o)$ \Comment{record clean activations}
\State $v \gets \mathcal{C}[n, t]$ \Comment{clean activation at $(n,t)$}
\State $f \gets \lambda x:\ v$ \Comment{patch hook returns the clean value}
\State $y' \gets M.\textsc{run\_with\_hooks}(o',\ \text{hooks}=\{(n,t): f\})$
\State \textbf{return} effect of restoring $(n,t)$ on the model output $y'$
\end{algorithmic}
\end{algorithm}

\section{Evaluation}
\label{sec:eval}

We evaluate three claims: that the interface covers world-model families through capability typing
(Section~\ref{sec:coverage}); that analyses written once recover meaningful structure on a real
model (Section~\ref{sec:casestudy}); and that the hook layer is cheap enough for always-on use
(Section~\ref{sec:overhead}).

\subsection{Coverage Through Capability Typing}
\label{sec:coverage}

Table~\ref{tab:coverage} lists the families currently implemented and the optional heads each
declares through its capability descriptor. The required core is identical across every family, and
the families differ only in which optional heads they expose, which is exactly the variation the
capability typing is designed to absorb. A reinforcement-learning model such as DreamerV3 populates
the full set of heads, a planning model such as PlaNet exposes a decoder and a reward head but no
actor, and a self-supervised model such as I-JEPA exposes none. Integrating a new backend therefore
reduces to implementing the four required methods and declaring capabilities, after which the entire
analysis library, including probing, activation patching, sparse autoencoders, and surprise
analysis, applies without modification. No analysis in the library contains architecture-specific
code.

\begin{table}[t]
\centering
\caption{Implemented backends and the capabilities each declares. The required core (encode,
transition, initial state, sample) is identical across every family; the families differ only in
which optional heads they expose. The full analysis suite applies to all of them without
modification.}
\label{tab:coverage}
\small
\begin{tabular}{lll}
\toprule
Family & Latent / dynamics & Declared optional heads \\
\midrule
DreamerV1 & continuous RSSM        & decode, reward, actor, critic \\
DreamerV2 & categorical RSSM       & decode, reward, continue, actor, critic \\
DreamerV3 & categorical RSSM       & decode, reward, continue, actor, critic \\
PlaNet    & continuous RSSM        & decode, reward \\
TD-MPC2   & joint embedding        & reward, actor, critic \\
IRIS      & transformer / codebook & decode, reward, continue \\
Decision Transformer & transformer & actor \\
I-JEPA    & joint embedding        & none (encoder \(+\) predictor) \\
Ha-Schmidhuber & VAE \(+\) MDN-RNN & decode, actor \\
\bottomrule
\end{tabular}
\end{table}

\subsection{Case Study: Layer-Resolved Structure in I-JEPA}
\label{sec:casestudy}

\begin{figure}[t]
\centering
\begin{tikzpicture}
\begin{axis}[
  width=7.6cm, height=4.2cm,
  ybar, bar width=11pt,
  enlarge x limits=0.18,
  ymin=-0.20, ymax=0.22,
  ylabel={Spearman ($\rho$)},
  xlabel={Predictor layer},
  symbolic x coords={0,1,2,3},
  xtick=data,
  ymajorgrids, grid style={gray!20},
  error bars/y dir=both, error bars/y explicit,
  axis line style={gray!60}, tick label style={font=\footnotesize},
  label style={font=\footnotesize},
  every axis plot/.append style={fill=wmlteal!60, draw=wmlteal}
]
\addplot+[error bars/.cd, y dir=both, y explicit] coordinates {
  (0, 0.001) +- (0,0.036)
  (1, -0.106) +- (0,0.036)
  (2, 0.134) +- (0,0.049)
  (3, -0.043) +- (0,0.038)
};
\draw[gray, dashed] (axis cs:0,0) -- (axis cs:3,0);
\end{axis}
\end{tikzpicture}
\caption{Attribution and attention agree only weakly inside the I-JEPA predictor. Spearman rank
correlation between integrated-gradients patch attribution and attention weight, per layer
(mean over the evaluation set, error bars are one standard error). Layer~2 shows modest positive
alignment, while layers~0, 1, and~3 are indistinguishable from zero or negative, indicating that
high-attention patches are not reliably the causally important ones.}
\label{fig:layersweep}
\end{figure}
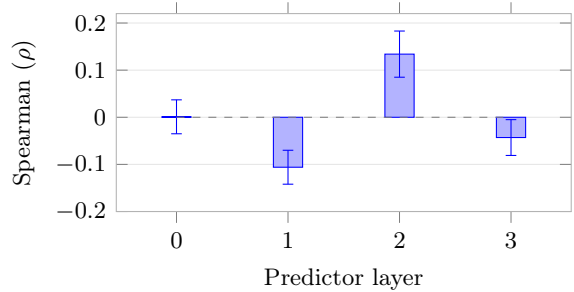

To show that the substrate recovers nontrivial structure, we apply the unmodified analysis suite to
the predictor of an I-JEPA model, itself a vision transformer \cite{dosovitskiy2021vit}, and ask, at
each layer, whether the context patches a head attends
to are the patches that causally determine its prediction. Context-patch importance is computed two
ways through the same hook interface. Integrated gradients \cite{sundararajan2017ig} integrate the
prediction gradient along a path from a mean-embedding baseline to the true patch embeddings over
fifty steps, scoring each context patch by its contribution to the predictor reconstruction loss
measured against the exponential-moving-average target encoder. Attention importance reads the
predictor self-attention weights directly from the cached \code{attn.hook\_pattern} activation,
taking the row of the masked target query over the context keys, per head and averaged over heads. An
evaluator compares the two rankings with the top-$k$ Jaccard overlap
$J_k = \lvert\mathcal{T}_k^{\mathrm{attn}}\cap\mathcal{T}_k^{\mathrm{IG}}\rvert \big/ \lvert\mathcal{T}_k^{\mathrm{attn}}\cup\mathcal{T}_k^{\mathrm{IG}}\rvert$
over the top-$k$ patch sets and the Spearman rank correlation $\rho$ between the two importance
vectors, reporting $95\%$ confidence intervals over the evaluation set. Figure~\ref{fig:layersweep} reports
the Spearman correlation across predictor layers. Agreement is weak throughout and is statistically
indistinguishable from zero in three of the four layers, with only one layer showing modest positive
alignment. The most-attended context patches are therefore frequently not the most causally relevant
ones, which is consistent with the broader caution that attention weight is an unreliable explanation
\cite{jain2019attention}. The contribution here is methodological. The comparison runs entirely
through the cache and hook interface with no model-specific code, and the same procedure transfers to
any adapter that exposes attention. We develop the mechanistic account of the effect in companion
work.

\subsection{Overhead of the Hook Layer}
\label{sec:overhead}

For interpretability telemetry to run inside a training or control loop, the hook layer must be
cheap when inactive. We measure per-step latency of the I-JEPA adapter in three conditions: the bare
adapter with no instrumentation; the adapter wrapped by \code{HookedWorldModel} with hook points
mounted but no hooks installed; and a full \code{run\_with\_cache} that records every activation.
Table~\ref{tab:overhead} reports the results. Mounting the hook layer adds approximately 12\% to
per-step latency, which is acceptable for always-on use, whereas exhaustive caching is roughly
seven times slower and is intended for offline analysis rather than continuous telemetry.
\emph{Reported numbers are to be finalized with the evaluation hardware and the number of steps
averaged, with mean and standard deviation.}

\begin{table}[t]
\centering
\caption{Per-step latency of the hook layer on the I-JEPA adapter. Mounting hook points is cheap;
exhaustive caching is reserved for offline analysis. Hardware and averaging details to be finalized.}
\label{tab:overhead}
\small
\begin{tabular}{lrr}
\toprule
Condition & ms / step & Overhead \\
\midrule
Bare adapter                 & 202.5  & baseline \\
Hook points mounted, inactive & 227.4  & $+12.3\%$ \\
Full \code{run\_with\_cache}  & 1360.4 & $6.7\times$ \\
\bottomrule
\end{tabular}
\end{table}

\section{Related Work}
\label{sec:related}

\textbf{Mechanistic interpretability tooling.} TransformerLens \cite{nanda2022transformerlens}
established the hook-and-cache abstraction for language models and underpins a large body of
circuit-level analysis \cite{elhage2021mathematical,wang2023ioi,conmy2023acdc}. The methods that run
on such tooling, including activation and path patching \cite{meng2022rome,goldowskydill2023pathpatching},
attribution patching \cite{nanda2023attributionpatching}, causal scrubbing \cite{chan2022causalscrubbing},
and the logit lens \cite{nostalgebraist2020logitlens}, share the read-and-edit interface we adopt.
NNsight \cite{fiottokaufman2024nnsight} exposes model internals for remote and local intervention,
Captum \cite{kokhlikyan2020captum} provides attribution primitives, and Penzai \cite{johnson2024penzai}
offers structural visualization and editing. These target generic networks or transformer language
models and do not represent world-model semantics. WorldModelLens centers the abstraction on the
world-model interface itself, including actions, dynamics rollouts, imagination, and intervention
replay, and treats reinforcement-learning and self-supervised models uniformly through capability
typing.

\textbf{World models.} The families we support span recurrent latent dynamics
\cite{ha2018worldmodels,hafner2019planet,hafner2020dreamer,hafner2021dreamerv2,hafner2023dreamerv3},
search-based latent models \cite{schrittwieser2020muzero,hansen2024tdmpc2}, transformer-token and
sequence models \cite{micheli2023iris,chen2021decisiontransformer,janner2021trajectory,bruce2024genie},
and joint-embedding predictive architectures grounded in self-supervised representation learning
\cite{bengio2013representation,assran2023ijepa,bardes2024vjepa,lecun2022path,he2022mae,grill2020byol,chen2020simclr}. World
foundation models extend the same ideas to video at scale \cite{nvidia2025cosmos}. These are the
targets of our analysis rather than competing tools, and no shared interpretability substrate
previously spanned them.

\textbf{Interpretability methods.} The analyses we make portable are drawn from work on probing
\cite{alain2016probes,belinkov2022probing,hewitt2019designing,li2023othello}, activation patching and causal
localization \cite{meng2022rome,wang2023ioi,goldowskydill2023pathpatching,vig2020causal}, sparse autoencoders and
dictionary learning \cite{olshausen1996sparse,cunningham2024saes,bricken2023monosemanticity,gao2024topk,rajamanoharan2024gated},
gradient and perturbation attribution
\cite{sundararajan2017ig,smilkov2017smoothgrad,selvaraju2017gradcam,lundberg2017shap,samek2017aopc},
representational similarity \cite{kornblith2019cka,raghu2017svcca}, disentanglement
\cite{chen2018mig,eastwood2018dci,kumar2018sap,higgins2017betavae}, and out-of-distribution detection
\cite{lee2018mahalanobis}. Our contribution is not these methods but a substrate that lets a single
implementation of each apply across world-model architectures.

\section{Limitations and Roadmap}
\label{sec:limitations}

WorldModelLens is alpha software, and several limitations bound the present claims. Our deep
empirical demonstration is on a single family, I-JEPA; the case for cross-architecture portability
rests on the shared interface and the implemented adapter coverage rather than on completed
cross-family studies, which are in progress for Dreamer at full scale, V-JEPA
\cite{bardes2024vjepa}, and Cosmos \cite{nvidia2025cosmos}. Some adapters provide faithful
re-implementations of an architecture rather than loading the original published checkpoints, and we
mark in the documentation which adapters load released weights. The overhead measurements in
Section~\ref{sec:overhead} are to be finalized with full hardware details and variance across runs.
We see closing these gaps, in particular reporting the same analysis across multiple loaded
checkpoints, as the primary path from this introduction to a full cross-architecture study.

\section{Conclusion}
\label{sec:conclusion}

We argued that the fragmentation of world-model interpretability is a property of the tooling rather
than of the models, and that a small capability-typed interface captures the structure shared across
latent-recurrent, transformer-token, and joint-embedding world models. WorldModelLens realizes this
interface and mounts a single hook, cache, rollout, and intervention-replay layer over it, so that
probing, patching, sparse autoencoders, and surprise analysis are written once and apply to every
backend. We demonstrated the substrate end-to-end on I-JEPA, reported adapter coverage across six
families, and showed that the hook layer is cheap enough for always-on telemetry. We hope the
substrate lowers the cost of interpretability research on world models and makes findings comparable
across the architectures that the field is rapidly producing.


{\small
\begin{thebibliography}{99}

\bibitem{ha2018worldmodels} D.~Ha and J.~Schmidhuber. Recurrent World Models Facilitate Policy Evolution. In \emph{NeurIPS}, 2018.
\bibitem{hafner2019planet} D.~Hafner, T.~Lillicrap, I.~Fischer, R.~Villegas, D.~Ha, H.~Lee, and J.~Davidson. Learning Latent Dynamics for Planning from Pixels. In \emph{ICML}, 2019.
\bibitem{hafner2020dreamer} D.~Hafner, T.~Lillicrap, J.~Ba, and M.~Norouzi. Dream to Control: Learning Behaviors by Latent Imagination. In \emph{ICLR}, 2020.
\bibitem{hafner2021dreamerv2} D.~Hafner, T.~Lillicrap, M.~Norouzi, and J.~Ba. Mastering Atari with Discrete World Models. In \emph{ICLR}, 2021.
\bibitem{hafner2023dreamerv3} D.~Hafner, J.~Pasukonis, J.~Ba, and T.~Lillicrap. Mastering Diverse Domains through World Models. \emph{arXiv:2301.04104}, 2023.
\bibitem{micheli2023iris} V.~Micheli, E.~Alonso, and F.~Fleuret. Transformers are Sample-Efficient World Models. In \emph{ICLR}, 2023.
\bibitem{chen2021decisiontransformer} L.~Chen, K.~Lu, A.~Rajeswaran, K.~Lee, A.~Grover, M.~Laskin, P.~Abbeel, A.~Srinivas, and I.~Mordatch. Decision Transformer: Reinforcement Learning via Sequence Modeling. In \emph{NeurIPS}, 2021.
\bibitem{assran2023ijepa} M.~Assran, Q.~Duval, I.~Misra, P.~Bojanowski, P.~Vincent, M.~Rabbat, Y.~LeCun, and N.~Ballas. Self-Supervised Learning from Images with a Joint-Embedding Predictive Architecture. In \emph{CVPR}, 2023.
\bibitem{hansen2024tdmpc2} N.~Hansen, H.~Su, and X.~Wang. TD-MPC2: Scalable, Robust World Models for Continuous Control. In \emph{ICLR}, 2024.
\bibitem{bardes2024vjepa} A.~Bardes, Q.~Garrido, J.~Ponce, X.~Chen, M.~Rabbat, Y.~LeCun, M.~Assran, and N.~Ballas. V-JEPA: Latent Video Prediction for Visual Representation Learning. \emph{arXiv:2404.08471}, 2024.
\bibitem{nvidia2025cosmos} NVIDIA. Cosmos World Foundation Model Platform for Physical AI. \emph{arXiv:2501.03575}, 2025.
\bibitem{nanda2022transformerlens} N.~Nanda and J.~Bloom. TransformerLens: A Library for Mechanistic Interpretability of Generative Language Models. \url{https://github.com/TransformerLensOrg/TransformerLens}, 2022.
\bibitem{elhage2021mathematical} N.~Elhage, N.~Nanda, C.~Olsson, T.~Henighan, et~al. A Mathematical Framework for Transformer Circuits. \emph{Transformer Circuits Thread}, 2021.
\bibitem{wang2023ioi} K.~Wang, A.~Variengien, A.~Conmy, B.~Shlegeris, and J.~Steinhardt. Interpretability in the Wild: A Circuit for Indirect Object Identification in GPT-2 Small. In \emph{ICLR}, 2023.
\bibitem{meng2022rome} K.~Meng, D.~Bau, A.~Andonian, and Y.~Belinkov. Locating and Editing Factual Associations in GPT. In \emph{NeurIPS}, 2022.
\bibitem{goldowskydill2023pathpatching} N.~Goldowsky-Dill, C.~MacLeod, L.~Sato, and A.~Arora. Localizing Model Behavior with Path Patching. \emph{arXiv:2304.05969}, 2023.
\bibitem{cunningham2024saes} H.~Cunningham, A.~Ewart, L.~Riggs, R.~Huben, and L.~Sharkey. Sparse Autoencoders Find Highly Interpretable Features in Language Models. In \emph{ICLR}, 2024.
\bibitem{bricken2023monosemanticity} T.~Bricken, A.~Templeton, J.~Batson, et~al. Towards Monosemanticity: Decomposing Language Models with Dictionary Learning. \emph{Transformer Circuits Thread}, 2023.
\bibitem{alain2016probes} G.~Alain and Y.~Bengio. Understanding Intermediate Layers Using Linear Classifier Probes. \emph{arXiv:1610.01644}, 2016.
\bibitem{belinkov2022probing} Y.~Belinkov. Probing Classifiers: Promises, Shortcomings, and Advances. \emph{Computational Linguistics}, 48(1), 2022.
\bibitem{kokhlikyan2020captum} N.~Kokhlikyan, V.~Miglani, M.~Martin, et~al. Captum: A Unified and Generic Model Interpretability Library for PyTorch. \emph{arXiv:2009.07896}, 2020.
\bibitem{fiottokaufman2024nnsight} J.~Fiotto-Kaufman, A.~R.~Loftus, E.~Todd, et~al. NNsight and NDIF: Democratizing Access to Foundation Model Internals. \emph{arXiv:2407.14561}, 2024.
\bibitem{johnson2024penzai} D.~D.~Johnson. Penzai and Treescope: Tools for Visualizing and Manipulating Neural Networks. \url{https://github.com/google-deepmind/penzai}, 2024.
\bibitem{sundararajan2017ig} M.~Sundararajan, A.~Taly, and Q.~Yan. Axiomatic Attribution for Deep Networks. In \emph{ICML}, 2017.
\bibitem{jain2019attention} S.~Jain and B.~C.~Wallace. Attention is not Explanation. In \emph{NAACL}, 2019.
\bibitem{kornblith2019cka} S.~Kornblith, M.~Norouzi, H.~Lee, and G.~Hinton. Similarity of Neural Network Representations Revisited. In \emph{ICML}, 2019.
\bibitem{chen2018mig} R.~T.~Q.~Chen, X.~Li, R.~Grosse, and D.~Duvenaud. Isolating Sources of Disentanglement in Variational Autoencoders. In \emph{NeurIPS}, 2018.
\bibitem{eastwood2018dci} C.~Eastwood and C.~K.~I.~Williams. A Framework for the Quantitative Evaluation of Disentangled Representations. In \emph{ICLR}, 2018.
\bibitem{kumar2018sap} A.~Kumar, P.~Sattigeri, and A.~Balakrishnan. Variational Inference of Disentangled Latent Concepts from Unlabeled Observations. In \emph{ICLR}, 2018.
\bibitem{higgins2017betavae} I.~Higgins, L.~Matthey, A.~Pal, et~al. beta-VAE: Learning Basic Visual Concepts with a Constrained Variational Framework. In \emph{ICLR}, 2017.
\bibitem{jang2017gumbel} E.~Jang, S.~Gu, and B.~Poole. Categorical Reparameterization with Gumbel-Softmax. In \emph{ICLR}, 2017.
\bibitem{maddison2017concrete} C.~J.~Maddison, A.~Mnih, and Y.~W.~Teh. The Concrete Distribution: A Continuous Relaxation of Discrete Random Variables. In \emph{ICLR}, 2017.
\bibitem{schrittwieser2020muzero} J.~Schrittwieser, I.~Antonoglou, T.~Hubert, et~al. Mastering Atari, Go, Chess and Shogi by Planning with a Learned Model. \emph{Nature}, 588, 2020.
\bibitem{janner2021trajectory} M.~Janner, Q.~Li, and S.~Levine. Offline Reinforcement Learning as One Big Sequence Modeling Problem. In \emph{NeurIPS}, 2021.
\bibitem{bruce2024genie} J.~Bruce, M.~Dennis, A.~Edwards, et~al. Genie: Generative Interactive Environments. In \emph{ICML}, 2024.
\bibitem{lecun2022path} Y.~LeCun. A Path Towards Autonomous Machine Intelligence. \emph{OpenReview}, 2022.
\bibitem{he2022mae} K.~He, X.~Chen, S.~Xie, Y.~Li, P.~Doll{\'a}r, and R.~Girshick. Masked Autoencoders Are Scalable Vision Learners. In \emph{CVPR}, 2022.
\bibitem{grill2020byol} J.-B.~Grill, F.~Strub, F.~Altch{\'e}, et~al. Bootstrap Your Own Latent: A New Approach to Self-Supervised Learning. In \emph{NeurIPS}, 2020.
\bibitem{chen2020simclr} T.~Chen, S.~Kornblith, M.~Norouzi, and G.~Hinton. A Simple Framework for Contrastive Learning of Visual Representations. In \emph{ICML}, 2020.
\bibitem{caron2021dino} M.~Caron, H.~Touvron, I.~Misra, et~al. Emerging Properties in Self-Supervised Vision Transformers. In \emph{ICCV}, 2021.
\bibitem{radford2021clip} A.~Radford, J.~W.~Kim, C.~Hallacy, et~al. Learning Transferable Visual Models from Natural Language Supervision. In \emph{ICML}, 2021.
\bibitem{olshausen1996sparse} B.~A.~Olshausen and D.~J.~Field. Emergence of Simple-Cell Receptive Field Properties by Learning a Sparse Code for Natural Images. \emph{Nature}, 381, 1996.
\bibitem{gao2024topk} L.~Gao, T.~Dupr{\'e} la Tour, H.~Tillman, et~al. Scaling and Evaluating Sparse Autoencoders. \emph{arXiv:2406.04093}, 2024.
\bibitem{rajamanoharan2024gated} S.~Rajamanoharan, A.~Conmy, L.~Smith, et~al. Improving Dictionary Learning with Gated Sparse Autoencoders. \emph{arXiv:2404.16014}, 2024.
\bibitem{conmy2023acdc} A.~Conmy, A.~Mavor-Parker, A.~Lynch, S.~Heimersheim, and A.~Garriga-Alonso. Towards Automated Circuit Discovery for Mechanistic Interpretability. In \emph{NeurIPS}, 2023.
\bibitem{nanda2023attributionpatching} N.~Nanda. Attribution Patching: Activation Patching at Industrial Scale. \url{https://neelnanda.io/attribution-patching}, 2023.
\bibitem{chan2022causalscrubbing} L.~Chan, A.~Garriga-Alonso, N.~Goldowsky-Dill, et~al. Causal Scrubbing: A Method for Rigorously Testing Interpretability Hypotheses. \emph{Alignment Forum}, 2022.
\bibitem{nostalgebraist2020logitlens} nostalgebraist. Interpreting GPT: The Logit Lens. \emph{LessWrong}, 2020.
\bibitem{smilkov2017smoothgrad} D.~Smilkov, N.~Thorat, B.~Kim, F.~Vi{\'e}gas, and M.~Wattenberg. SmoothGrad: Removing Noise by Adding Noise. \emph{arXiv:1706.03825}, 2017.
\bibitem{selvaraju2017gradcam} R.~R.~Selvaraju, M.~Cogswell, A.~Das, R.~Vedantam, D.~Parikh, and D.~Batra. Grad-CAM: Visual Explanations from Deep Networks via Gradient-Based Localization. In \emph{ICCV}, 2017.
\bibitem{lundberg2017shap} S.~M.~Lundberg and S.-I.~Lee. A Unified Approach to Interpreting Model Predictions. In \emph{NeurIPS}, 2017.
\bibitem{samek2017aopc} W.~Samek, A.~Binder, G.~Montavon, S.~Lapuschkin, and K.-R.~M{\"u}ller. Evaluating the Visualization of What a Deep Neural Network Has Learned. \emph{IEEE TNNLS}, 28(11), 2017.
\bibitem{raghu2017svcca} M.~Raghu, J.~Gilmer, J.~Yosinski, and J.~Sohl-Dickstein. SVCCA: Singular Vector Canonical Correlation Analysis for Deep Learning Dynamics and Interpretability. In \emph{NeurIPS}, 2017.
\bibitem{hewitt2019designing} J.~Hewitt and P.~Liang. Designing and Interpreting Probes with Control Tasks. In \emph{EMNLP}, 2019.
\bibitem{lee2018mahalanobis} K.~Lee, K.~Lee, H.~Lee, and J.~Shin. A Simple Unified Framework for Detecting Out-of-Distribution Samples and Adversarial Attacks. In \emph{NeurIPS}, 2018.
\bibitem{vaswani2017attention} A.~Vaswani, N.~Shazeer, N.~Parmar, et~al. Attention Is All You Need. In \emph{NeurIPS}, 2017.
\bibitem{dosovitskiy2021vit} A.~Dosovitskiy, L.~Beyer, A.~Kolesnikov, et~al. An Image Is Worth 16x16 Words: Transformers for Image Recognition at Scale. In \emph{ICLR}, 2021.
\bibitem{kingma2014vae} D.~P.~Kingma and M.~Welling. Auto-Encoding Variational Bayes. In \emph{ICLR}, 2014.
\bibitem{sutton2018rl} R.~S.~Sutton and A.~G.~Barto. \emph{Reinforcement Learning: An Introduction}. MIT Press, 2nd edition, 2018.
\bibitem{oquab2024dinov2} M.~Oquab, T.~Darcet, T.~Moutakanni, et~al. DINOv2: Learning Robust Visual Features without Supervision. \emph{Transactions on Machine Learning Research}, 2024.
\bibitem{bengio2013representation} Y.~Bengio, A.~Courville, and P.~Vincent. Representation Learning: A Review and New Perspectives. \emph{IEEE Transactions on Pattern Analysis and Machine Intelligence}, 35(8), 2013.
\bibitem{li2023othello} K.~Li, A.~K.~Hopkins, D.~Bau, F.~Vi{\'e}gas, H.~Pfister, and M.~Wattenberg. Emergent World Representations: Exploring a Sequence Model Trained on a Synthetic Task. In \emph{ICLR}, 2023.
\bibitem{kim2018tcav} B.~Kim, M.~Wattenberg, J.~Gilmer, C.~Cai, J.~Wexler, F.~Vi{\'e}gas, and R.~Sayres. Interpretability Beyond Feature Attribution: Quantitative Testing with Concept Activation Vectors (TCAV). In \emph{ICML}, 2018.
\bibitem{geiger2021causal} A.~Geiger, H.~Lu, T.~Icard, and C.~Potts. Causal Abstractions of Neural Networks. In \emph{NeurIPS}, 2021.
\bibitem{templeton2024scaling} A.~Templeton, T.~Conerly, J.~Marcus, et~al. Scaling Monosemanticity: Extracting Interpretable Features from Claude 3 Sonnet. \emph{Transformer Circuits Thread}, 2024.
\bibitem{vig2020causal} J.~Vig, S.~Gehrmann, Y.~Belinkov, S.~Qian, D.~Nevo, Y.~Singer, and S.~Shieber. Investigating Gender Bias in Language Models Using Causal Mediation Analysis. In \emph{NeurIPS}, 2020.

\end{thebibliography}
}

\end{document}